\documentclass[runningheads]{llncs}
\usepackage{comment}
\usepackage{xcolor}%
\usepackage{textcomp}%
\usepackage{manyfoot}%
\usepackage{booktabs}%
\usepackage{algorithm}%
\usepackage{algorithmicx}%
\usepackage{algpseudocode}%
\usepackage{listings}%
\usepackage{tikz}
\usepackage{color}
\usetikzlibrary{chains,matrix,scopes,shapes.geometric}
\usepackage{hyperref}
\usepackage{footnote}
\usepackage{todonotes}
\usepackage[utf8]{inputenc}

\usepackage[T1]{fontenc}
\usepackage{graphicx}
\begin{document}
\title{Automated Creation of the Legal Knowledge Graph Addressing Legislation on Violence Against Women: Resource, Methodology and Lessons Learned}
%
\titlerunning{Automated Creation of the Legal Knowledge Graph}
%
\author{Claudia d'Amato\inst{1}\orcidID{0000-0002-3385-987X} \and
Giuseppe Rubini\inst{1} \and
Didio Franceso\inst{1}\and
Francioso Donato\inst{1}\and
Fatima Zahra Amara\inst{1}\orcidID{0000-0001-8463-0330}\and
Nicola Fanizzi\inst{1}\orcidID{0000-0001-5319-7933}}
\authorrunning{C. d'Amato et al.}
%
\institute{Computer Science Department, University of Bari Aldo Moro, Italy
\email{\{claudia.damato, fatima.amara, nicola.fanizzi\}@uniba.it}\\
\email{\{g.rubini13, f.didio2, d.francioso7\}@studenti.uniba.it}}
\maketitle 
\begin{abstract}
Legal decision-making process requires the availability of comprehensive and detailed legislative background knowledge and 
up-to-date information on legal cases and related sentences/decisions. Legal Knowledge Graphs (KGs) would be a valuable tool to facilitate access to legal information, to be queried and exploited for the purpose, and to enable advanced reasoning and machine learning applications. 
Indeed, legal KGs may act as 
knowledge intensive component to be 
used by predictive machine learning solutions supporting the decision process of the legal expert. Nevertheless, a few KGs can be found in the legal domain. To fill this gap, 
we developed a legal KG targeting legal cases of violence against women, along with clear adopted methodologies. 
Specifically, the paper introduces two complementary 
approaches for automated legal KG construction; a systematic bottom-up approach, customized for the legal domain, and a new solution leveraging Large Language Models. Starting from legal sentences publicly available from the European Court of Justice, 
the solutions integrate structured data extraction, ontology development, and semantic enrichment to produce KGs tailored for legal cases involving violence against women. After analyzing and comparing the results of the two approaches, the developed KGs are validated via suitable competency questions. 
The obtained KG may be impactful for multiple purposes: can improve the accessibility to legal information both to humans and machine, can enable complex queries and may constitute an important knowledge component to be possibly exploited by machine learning tools tailored for predictive justice.\\ 
\textbf{Resource type:} Knowledge Graphs\\
\textbf{License:} Creative Commons Attribution 4.0 International \\
\textbf{DOI:} \url{https://doi.org/10.5281/zenodo.15270173} \\
\textbf{URL:} \url{https://lod-cloud.net/dataset/PREJUST4WOMAN\_PROJECT}
\end{abstract}

\section{Introduction}
In recent years, the legal domain has been interested by an increasing usage of Artificial Intelligence (AI) solutions \cite{diPorto2024,li2024construction} to effectively manage and mine legal consultation and decision process. 
Despite 
advancement in legal technologies~\cite{Westermann19}, a significant gap exists in the availability of structured and easy to query resources tailored to the legal domain such as Knowledge Graphs (KGs)~\cite{hogan2022knowledge}. Existing KGs in the legal domain often fail to address nuanced and domain-specific challenges, including the need for interoperability, semantic richness, and adaptability to predictive justice applications \cite{changyue_wang__2024,xuran_wang__2024}.
A legal KG needs to be able to cope with the specific and intricate writing style of laws and sentences, and 
keeping  track of the evolution of laws, including exceptions and applicability.

We specifically focus on the widespread and urgent issue of violence against women\footnote{\url{https://www.who.int/news-room/fact-sheets/detail/violence-against-women}}  
that leads to physical and mental health problems, economic losses and social instability~\cite{dawa2022violence} and that impacts societies around the world. Manifesting in physical and emotional forms, it constitutes a grave violation of human rights, threatening the safety, dignity, and well-being of countless women. 
%
Actually, a multifaceted approach involving governments, civil society and individuals is essential to counter this pervasive problem and promote healthier and more equitable societies. 
Our work aims to contribute to this effort by developing advanced instruments and knowledge components to assist in the legal handling of cases related to violence against women. 
%
%
This 
goal requires robust 
tools 
capable of supporting legislative interpretation, judicial decision-making, policy formulation 
and access to structured resources, in the form of KGs, tailored to cases of gender-based violence. 
%
%
%
For this purpose, 
we introduce a novel 
Legal KG, specifically designed to address legislation and judicial cases related to violence against women, jointly with methodologies adopted for its development. 


The 
Legal KG is automatically derived from the jurisprudence of sentences of the European Court of Human Rights (ECHR\footnote{\url{ https://www.echr.coe.int/}}), 
targeting cases of gender-based violence, and it 
adheres to the FAIR (\textit{Findable}, \textit{Accessible}, \textit{Interoperable}, \textit{Reusable}) data principles. 
%
The resource is built upon two 
methodologies: a customization to the legal domain of the general KG development process
~\cite{Tamaauskaite2022} (that is a bottom-up solution 
combining data extraction, ontology development, and semantic enrichment via 
alignment to domain-specific ontologies), 
and a new approach for automated KG generation, grounded on the exploitation of Large Language Models (LLMs) as the recent advancements have opened up new possibilities for automating and improving various phases of ontology development \cite{li2025large}, and 
customized to the legal domain. 
The two methodologies resulted complementary: the former providing more precise outcomes but more time-consuming, the latter more scalable but limited in accuracy.
The obtained resource has been also validated via suitable competency questions~\cite{Monfardini2023}. 

Besides the Legal KG being the first result addressing legislation,  particularly concerning violence against women, this resource may 
act as proxy 
for broader applications. In the context of predictive justice, sentence prediction may be regarded as a link prediction problem on (a subset of) the Legal KG. This approach may be generalized to other domains in addition to the one of violence against women, provided the availability of legal KGs 
generated  
by adopting the proposed methodologies. 

The remainder of this paper is structured as follows: 
Section~\ref{related} review the 
state of the art 
on legal informatics, KGs and ontologies in law. 
Section~\ref{approach-bottomUp} presents a customization to the legal domain of the bottom-up KG development process 
and the obtained 
Legal KG. 
Section~\ref{approach-LLM} presents a new approach for automated KG generation, grounded on the exploitation of Large Language Models (LLMs) and 
customized to the legal domain, 
the obtained Legal KG.
Section~\ref{disc} discusses the results and 
the effectiveness of the adopted methodologies. Section~\ref{conc} conclude the paper 
and delineate future research directions. 

\section{Related Work}\label{related}
In this section, we survey the state of the art 
at the intersection of legal informatics, 
usage of KGs for the legal domain and development of ontologies and legal thesaurus initiatives. 
These studies explore methods for structuring legal data, applying automated reasoning, and extracting insights from legal texts. 

Legal knowledge extraction and presentation of 
the Joint Knowledge Enhancement Model (JKEM), which refines LLMs using a 
prefix-based strategy to improve knowledge extraction accuracy is proposed in~\cite{li2024construction}. 
Using this paradigm, the authors created the Chinese Legal Knowledge Graph (CLKG), which has 3480 knowledge triples but does not exploit any existing domain ontology. 
Similarly, 
in~\cite{zhao2024method} a judicial case KG centered mostly on event elements, employing a joint event extraction model to mine entity and event information from digital case records is presented. Using the Dgraph graph database for storage and GraphQL for querying, this approach facilitates 
retrieval, case analysis, and knowledge discovery but does not push on interlinking the resource with other existing resources. 
Analogously, 
in ~\cite{stavropoulou2020architecting} 
the ManyLaws platform is presented, 
a Big Open Legal Data (BOLD) platform that provides advanced and customizable access to legal information across the EU. The platform offers services such as legal corpus research, law comparison, legislative alignment analysis, and legislative timelines visualization. By leveraging data analytics, text mining, 
and interactive visualization, the platform enhances accessibility and understanding of legal data for a wide range of users but it does not focus on data interlinking and semantic querying. 
An approach for automatic legal document generation based on KGs is presented in~\cite{wei2024intelligent}. The method improves generation efficiency, quality, and content integrity 
compared to traditional rule-based approaches, contributing to 
increased automation and precision in legal document production but, it does not allow 
semantic query of legal documents. 

Focusing more on operational level, 
Francesconi et al.~\cite{francesconi2023patterns} presented a 
solution for legal compliance checking 
by exploiting ontologies.  Specifically, classes and property restrictions are used to model deontic norms. 
Legal reasoning is performed by decidable reasoners, and the framework is generalized through reusable patterns for modeling deontic norms and compliance verification. 
An RDF-based graph for representing and searching specific parts of legal documents, such as paragraphs and clauses, rather than retrieving entire documents is formulated in~\cite{oliveira2024rdf}. The graph, grounded in an ontological framework, models both the general structure of the legal system and the internal composition of legal documents. This enables the efficient retrieval of relevant legal frameworks for specific topics. Nevertheless, the resource comes from a significant manual effort with limited automation of its construction. 

As for 
KGs in the legal domain, a seminal example has been developed within 
the Lynx project~\cite{schneider2022lynx}, that focusing on GDPR regulation and contract compliance, 
developed a KG\footnote{\url{https://lynx-project.eu/doc/lkg/}} 
that structures document relationships and incorporates legal linguistic information. 
Inspired by the Lynx project, 
a pipeline 
for extracting information from documents produced by the \textit{Istituto Poligrafico Zecca dello Stato} in order to make them easy to query is formulated in~\cite{Anelli23}. The system consists of multiple  
modules, 
but shows limited re-use of existing domain ontologies. A KG for Vietnamese legal cases has been 
proposed in~\cite{Vuong23}. The KG is created from a data-source of Vietnamese laws and legal cases. 
As for~\cite{Anelli23}, 
a limited re-use of existing domain ontologies is done. This limitation also applies to the system presented in~\cite{Sovrano20}, that is intended to answer questions posed to search for information related to specific topics, concepts or entities. Nevertheless, this solution poses increased attention to the ontology engineering perspective of the extracted knowlege by adopting ontology design patterns. 

Ontologies and thesaurus for the legal field have been proposed~\cite{Filtz21} with the goal of  
standardizing legal terminology and facilitating interoperability across different legal systems. In the following the main initiatives are surveyed. 
\begin{description}
    \item \textbf{EuroVoc }\footnote{\url{https://op.europa.eu/en/web/eu-vocabularies/dataset/-/resource?uri=http://publications.europa.eu/resource/dataset/eurovoc}} is a multidomain, multilingual thesaurus provided by the \textit{Publications Office} of the European Union used to classify EU documents into categories to facilitate information searching. 
It is based on the \textit{Simple Knowledge Organization System}\footnote{\url{https://www.w3.org/TR/skos-reference/}} (SKOS) standard that is used to represent and organize concepts and vocabularies.
    \item \textbf{European Law Identifier (ELI)}\footnote{\url{https://op.europa.eu/en/web/eu-vocabularies/eli}, \url{https://eur-lex.europa.eu/eli-register/about.html}}: ELI is a standard created to identify legislative documents from European states by providing an ontology and metadata. It provides simplified access, exchange and reuse of legislation for legal professionals or citizens, and forms the basis for a representation of the Official Journals of the member states in the Semantic Web\footnote{Major progress towards transparency: a new European legislation identifier: \url{https://ec.europa.eu/commission/presscorner/detail/en/IP_12_1040}.}. 
    \item \textbf{European Case Law Identifier (ECLI)} \footnote{\url{https://e-justice.europa.eu/topics/legislation-and-case-law/european-case-law-identifier-ecli_en}, \url{https://eur-lex.europa.eu/content/help/eurlex-content/ecli.html}}: ECLI defines a  standard identifier  for European jurisprudence, together with a minimal set of metadata. The identifier provided by ECLI is intended to have an identification code whose structure is common among all member states. 
    This is divided into five parts separated by the colon as in the following example: \texttt{ECLI:CE:ECHR:2022:0210JUD007397516}.
\end{description}
ECLI provides a European system for the identification of case-law. ELI identifies legislative texts which have different and more complex characteristics. The two solutions are somehow complementary. In general, ECLI is preferable for legal decisions, while ELI is suitable for legislative texts.

To the best of our knowledge, our work represents the first resource targeting legal cases of violence against women, releasing a KG queryable via standard languages such as SPARQL, compliant with FAIR principle, and re-using existing domain ontologies and legal thesaurus. Our Legal KG addresses both the specificity of European legislation and the broader need for advanced tools to support legal reasoning and decision-making.

\section{Legal Knowledge Graph Construction Using a Bottom-Up Approach}\label{approach-bottomUp}
The advancements in KG engineering have significantly transformed various domains including legal studies, healthcare, public administration, and education. 
In the legal domain, KGs have been instrumental in organizing and analyzing complex legislative texts, case law, and regulatory frameworks, enabling enhanced legal reasoning and predictive justice. 
They also improve data integration, support intelligent decision making, and facilitate complex queries across disparate data sets. 

A general methodology for the KG development 
has been proposed by Tamašauskaitė and Groth~\cite{Tamaauskaite2022}.
Two 
approaches are distinguished: 
\textit{Top-down}, when an ontology is defined and based on it, knowledge is extracted from the data; \textit{Bottom-up}, when knowledge is extracted from data and then an ontology is created to represent the extracted data. 

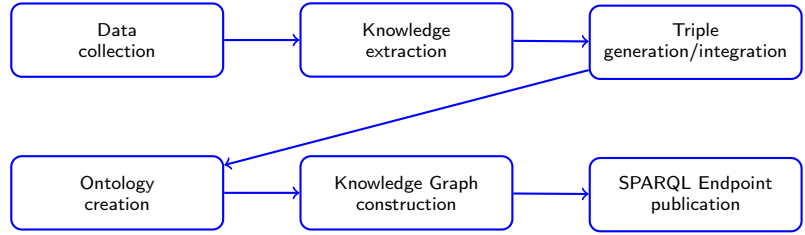
\begin{figure}[tb]
    \centering
\begin{tikzpicture}[every node/.style=draw, thick, text width=8em, align=center, minimum height=3em, rounded corners, font=\scriptsize\sffamily, draw=blue]
  \matrix [matrix of nodes, column sep=1cm, row sep=1cm, draw=none, ]
  {
    |(a)| {Data\\ collection}& |(b)| {Knowledge\\ extraction}& |(c)| {Triple\\ generation/integration}\\
    |(d)| {Ontology\\ creation}& |(e)| {Knowledge Graph\\ construction}& |(f)| {SPARQL Endpoint\\ publication}\\
  };
  { [start chain,every on chain/.style={join=by ->}]
    \chainin (a);
    \chainin (b);
    \chainin (c);
    \chainin (d);
    \chainin (e);
    \chainin (f);
  }
\end{tikzpicture}
    \caption{The adopted pipeline.}
    \label{fig3}
\end{figure}

In this section, we present a customization of the bottom-up approach to the legal domain, specifically for creating 
a KG for cases of violence against women. 
The methodology 
consists of six key steps 
encompassing data collection, 
knowledge extraction, knowledge processing and triple generation, ontology creation, KG construction, and maintenance, as summarized in Fig.~\ref{fig3}, where we specifically highlight our goal of making the resulting KG accessible via a SPARQL endpoint. 
In the following, we describe the application of each step to the legal domain, specifically focusing on cases of violence against women. 

The system, implementing 
the entire pipeline, has been made freely available in the GitHub repository\footnote{\url{https://github.com/PeppeRubini/EVA-KG}}, as Phyton project, jointly with  documentation. When needed, we also provide implementation details in the following description and provide specific pointers for them within the GitHub project. 

Regarding \textbf{data collection}, ECHR judgments and decisions, accessible through the official website\footnote{\url{https://hudoc.echr.coe.int/}} have been collected and referenced in a compiled document\footnote{\url{https://github.com/PeppeRubini/EVA-KG/blob/main/data/mapping_doc_link.xlsx}}. 
Specifically, 
$73$ judgments and decisions (in English\footnote{ \url{https://hudoc.echr.coe.int/eng}}) 
have been selected by experts in international law. 
For each 
judgement/decision, 
we specifically accessed 
tabs 
\texttt{View} and \texttt{Case Details} where 
\texttt{View} tab provides the full pdf document, while 
\texttt{Case Details} tab contains additional semi-structured information (in HTML format), such as publication details, importance level, and ECLI identifiers, in agreement with the 
specific format: \texttt{ECLI:EC:ECHR:year}. 
For example, \texttt{ECLI:EC:ECHR:2022:0210JUD007397516} denotes a judgment from February 10, 2022, with appeal number \texttt{73975/16}. 
Using Selenium 4 Library\footnote{\url{https://selenium.dev/}}, we set up a functionality\footnote{\url{https://github.com/PeppeRubini/EVA-KG/blob/main/src/echr_scraper.py}} for automatic download 
PDF and HTML files for each url sentence,  
and have them ready for the successive knowledge extraction step. Indeed, Selenium allows to control a web browser automatically and simulate human actions, such as clicking with the mouse on a specific item. 

The successive \textbf{Knowledge Extraction} and \textbf{Triple Generation} phases focus on processing sentence files 
for extracting corresponding triples. 
For this purpose, HTML files have been considered as more comprehensive. Processing of each sentence instantiates a \texttt{ECHRDocument} class, representing a generic ECHR document. 
The Beautiful Soup library\footnote{\url{https://beautiful-soup-4.readthedocs.io/en/latest/}} has been used to extract information, which includes converting defendant state names to Wikidata URIs and modifying the Importance Level values for consistency.
Given the $73$ selected judgments (65) and decisions (8), overall $10325$ triples have been extracted, counting $22$ distinct predicates and $5185$ distinct entities. 
The extracted triples are either serialized into a JSON file or 
processed via RDFLib library\footnote{\url{https://rdflib.readthedocs.io/en/stable/gettingstarted.html}}, 
that facilitates the creation of triples in standard RDF format, that can then be serialized into Turtle files. A further check for redundancy elimination has been performed, but no redundancy among the triples derived from each sentence has been found. Additionally, when combining the triples from all sentences, 
RDFLib automatically removes any duplicates found across different files.



\begin{figure}[h]
\centering
\includegraphics[width=15cm]{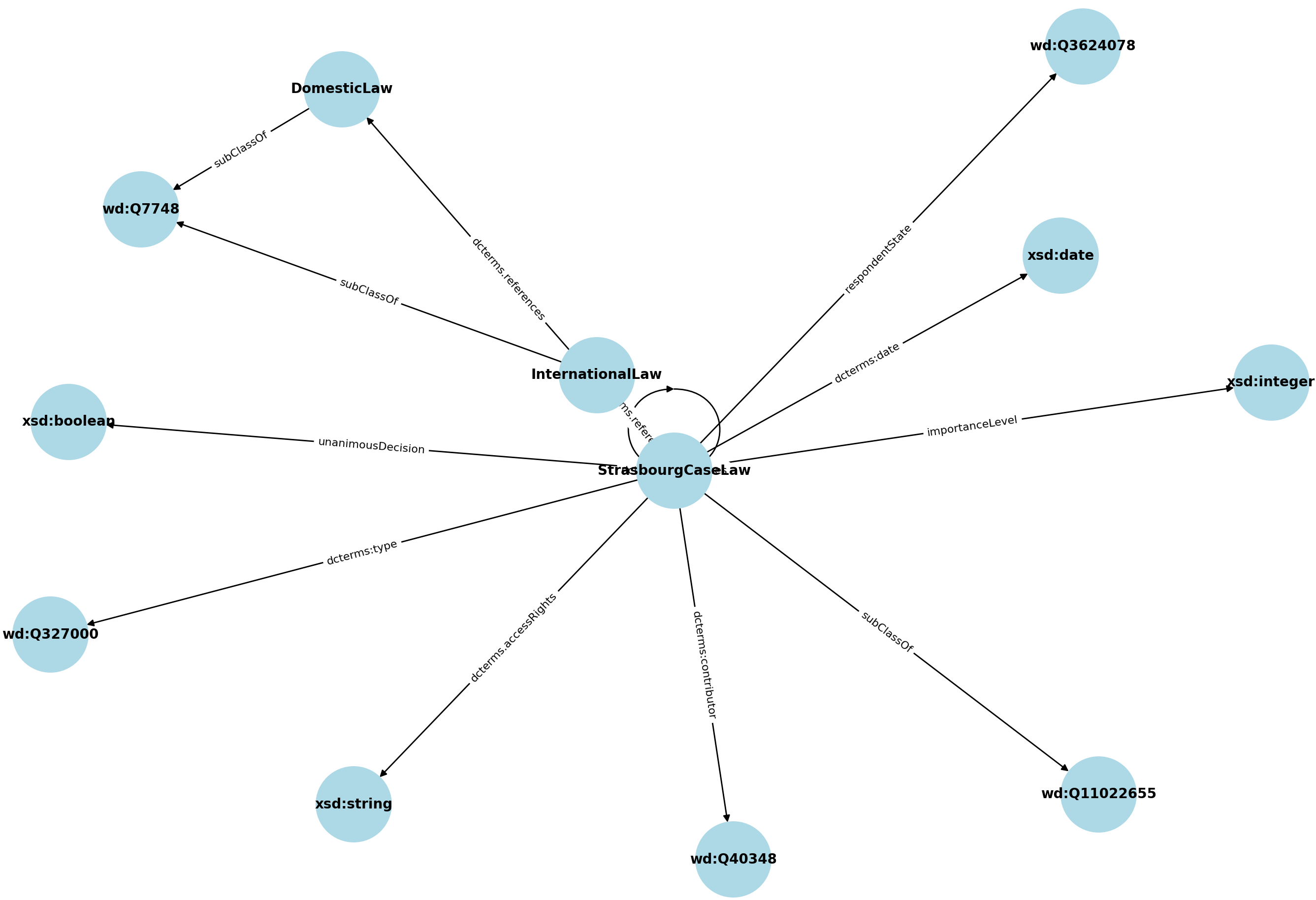}
\caption{Part of Knowledge Graph Constructed Using a Bottom-Up Approach.}
\label{KGapp1}
\end{figure}
As for the \textbf{Ontology Creation} and \textbf{KG construction}, in agreement with the ontology engineering best practices, we started with the formalization of competency questions\footnote{\url{https://protege.stanford.edu/publications/ontology\_development/ontology101-noy-mcguinness.html}} (CQs) that the ontology-based KG should address. The CQs adopted for this purpose are summarized in Tab.~\ref{tab:CQ1} (first column). 
For this phase, we also considered existing ontologies and vocabulary, 
such as ECLI, for possible reuse for 
representing ECHR pronouncements. However, not all CQs could have been answered using the existing resources (Tab.~\ref{tab:CQ1}, second column). 
Hence, new concepts/classes and properties have been defined. These new concepts include 
\texttt{DomesticLaw}, \texttt{InternationalLaw}, and \texttt{StrasbourgCaseLaw}. As for the properties, 
\texttt{applicationNumbers}, \texttt{importanceLevel}, \texttt{respondentStates}, \texttt{involvedArticles}, and \texttt{unanimousDecisionIndicators}, have been  defined. Furthermore, new individuals, represented by URI sentences, have been created for each ruling, instantiating 
the \texttt{StrasbourgCaseLaw} class. Figure \ref{KGapp1} illustrates a fragment of the hierarchical taxonomy of the legal concepts in the developed KG, such as \texttt{DomesticLaw} and \texttt{StrasbourgCaseLaw}, along with associated properties. All in all, the metadata 
from Case Details have been used to enrich the ECLI data and address most of the CQs. 
%
The final KG has been constructed by combining the triples from various judgments and linked to external resources including 
Wikidata for geographic
information representation. We also developed three independent visualization solutions by adopting: 
PyVis Library\footnote{\url{https://pypi.org/project/pyvis/}}, RDF Grapher\footnote{\url{https://www.ldf.fi/service/rdf-grapher}}, and Neo4j\footnote{\url{https://neo4j.com/}}, respectively.

In order to query the obtained KG, we proceeded with the \textbf{SPARQL Endpoint publication}. Specifically, the SPARQL endpoint has been implemented as a Flask\footnote{\url{https://flask.palletsprojects.com/en/3.0.x/}} web application, leveraging the flexibility and simplicity of the Flask framework for web services. Flask, serves as the web
framework for handling HTTP requests, routing, and rendering responses. SPARQL 1.1 engine has been used for executing the SPARQL queries. 
RDF data is loaded into the RDFLib graph at application startup. The graph supports multiple serialization formats, such as Turtle, RDF/XML, and N-Triples.
The application includes two main routes: the home page (\texttt{/}) with a form for submitting queries and a query endpoint (\texttt{/query}) that processes and displays the results. The interface uses HTML templates: \texttt{INDEX.html} for query submission and \texttt{RESULT.html} for displaying results. This setup provides a user-friendly way to interact with and query RDF data.
The SPARQL endpoint, along with the full implementation of the web application, is available on the GitHub repository\footnote{\url{https://github.com/khaoulafatima/PJ4W}}.

\begin{table}[bt]
\centering
 \caption{
 Competency Questions Adopted for the Bottom-Up Approach.}
\resizebox{\textwidth}{!}{%
\begin{tabular}{|p{10cm}|p{4cm}|} \hline 
\textbf{Question }& \textbf{Corresponding Term}\\ \hline 
What type of document are we dealing with?    & dcterms:type\\ \hline 
When is the document dated?    & dcterms:date\\ \hline 
Can I retrieve the information of a certain case given its identifier? & dcterms:isVersionOf \\ \hline 
Who represents the applicant?	& dcterms:contributor\\ \hline 
To which European state does the applicant belong?& respondentState\\ \hline 
What was the ruling?	& dcterms:abstract\\ \hline 
Was the ruling unanimous?	& unanimousDecision\\ \hline 
Which articles of the Convention were considered?& involveConventionArticle\\ \hline 
Which laws were considered to reach this conclusion?& dcterms:references\\ \hline
What is the importance of the ruling concerning future cases?&importanceLevel\\\hline
Is the document publicly accessible?	&dcterms:accessRights\\\hline
In which language is the document written?	&dcterms:language\\\hline
Where can I consult the document?	&dcterms:identifier\\\hline
    \end{tabular}
    } 
    \label{tab:CQ1}
\end{table}

The resulting KG, publicly available
\footnote{\url{https://github.com/PeppeRubini/EVA-KG/blob/main/KG.ttl}}, provides a semantically rich interconnected structure that enables advanced querying and analysis of ECHR decisions pertaining to gender-based violence. It leverages established semantic web standards and vocabularies, enabling interoperability with external resources and datasets.
The dataset has been published as part of the LOD cloud\footnote{The {PREJUST4WOMAN} dataset can be found at: \url{https://lod-cloud.net/dataset/PREJUST4WOMAN\_PROJECT}}, adhering to the FAIR principles. 

\section{Legal Knowledge Graph Construction Using Large Language Models}
\label{approach-LLM}
Recent advances in LLM have revolutionized NLP by providing powerful tools for understanding and generating text. 
Models such as GPT-4 and Mixtral provide capabilities that extend to ontology development, KG construction, and query answering, creating new opportunities for automating knowledge engineering tasks in specialized domains. 
The LLM methodology improves accuracy and efficiency in information extraction for KG production while overcoming the constraints of existing methods \cite{chen2024information}.\\
These models can augment traditional approaches by leveraging extensive training data to infer relationships, identify patterns, and construct meaningful representations of unstructured textual data, such as court rulings and legal documents.\\
Legal document analysis technology based on LLMs, with the semantic understanding capacity and contextual modeling ability, may effectively tackle these challenges and provide powerful auxiliary tools for lawyer \cite{ammar2024prediction}. This synthesis of LLM and KG enables better understanding and generation of structured information, addressing long-standing challenges in the field.
The development of KGs for legal texts presents unique challenges due to the complexity and specificity of the domain. This approach combines the capabilities of LLMs and traditional NLP techniques to address these challenges. LLMs offer advanced capabilities for ontology generation and data extraction, while NLP techniques provide structured processing of text to ensure precision and clarity. By integrating these methods into a unified pipeline, the proposed approach leverages the strengths of both to create accurate and comprehensive KGs that are aligned with domain requirements and validated through competency questions.

Effective prompt engineering is essential for leveraging LLMs to produce desired outputs\footnote{\url{https://www.diariodiunanalista.it/posts/guida-prompt-engineering/}}. This approach utilized two key techniques: \textit{zero-shot prompting}, suitable for simple tasks with clearly expressed requirements, and \textit{few-shot prompting}, which includes examples to guide the model in handling more complex tasks and reducing errors like hallucination.\\

The proposed methodology for generating KGs from legal rulings integrates advanced LLMs and traditional NLP techniques into a cohesive workflow. It involves two pipelines, one of which is an NLP-based pipeline\footnote{\url{https://github.com/Fra3005/PreJust4Womans}}. The LLM pipeline includes document preparation, Retrieval-Augmented Generation (RAG) development, ontology and KG creation, and competency question validation, while the NLP pipeline focuses on preprocessing, POS tagging, and triple extraction for constructing KGs.
\subsubsection{KG Development}
\begin{enumerate}
\item \textbf{LLM's Pipeline}\\
\textbf{- Document Preparation:} The case study focused on analyzing and constructing KG from legal judgments, considering two distinct types of input. The first type utilized the entire judgment (full-text), while the second involved a specific sub-part of the judgment, selected by domain experts. The aim was to assess the effectiveness of both input types. For preparation, automated functions were used to extract text from PDFs for the full-text input. However, due to the varying structure and complexity of each judgment, manual extraction was employed for the sub-part input. Once the text was extracted, it was formatted into a document, ready for the subsequent steps.\\
\textbf{- RAG's Creation:} RAG models have been employed to limit the output of LLMs to the context of each specific document. Although LLMs are trained on vast amounts of data, they can sometimes struggle with providing accurate responses to specialized tasks. To address this, RAGs enhance traditional LLMs by integrating a well-defined corpus of text \cite{lewis2020retrieval}. This enables LLMs to access two types of data when responding to user queries: parametric data, which refers to the model's training data, and nonparametric data, which consists of new, external data stored in a vector container. The LLM inspects this nonparametric data to generate precise answers.\\
For the creation of RAGs, the TogetherEmbedding method \footnote{\url{https://python.langchain.com/v0.2/docs/integrations/text\_embedding/together/}} from the LangChain library was used. BERT-M2 served as the embedding model for semantic search within the vector dataset, and FAISS \footnote{\url{https://python.langchain.com/v0.1/docs/integrations/vectorstores/faiss/}}, accessed through the langchain\_community library, was used to manage the vector dataset. RAGs were created for all five documents involved, which were then used as the contextual basis for the LLM to generate responses to various prompts.
\textbf{- Base's Ontology Creations:}
At this stage, the ontology, serving as a T-box for constructing the final KGs, was developed. Initially, GPT-4.o was employed to generate a foundational ontology, as Mixtral 8x22b proved incapable of creating a general schema for this topic. Using GPT-4.o resulted in a basic ontology containing core elements such as classes (e.g.,\textit{ Abuse and LegalCase}) and \textit{ObjectProperty}. Subsequently, the ontology was enriched with additional elements using Mixtral, which, despite explicit instructions in the prompts, tended to add extra components like \textit{DataProperty }and \textit{ObjectProperty} during the KG generation process.\\
To further refine the ontology, a zero-shot prompting technique was applied to domain-specific documents. Prompts instructed the model to expand the ontology by incorporating fundamental concepts from these documents. This process was repeated with several randomly selected documents. After multiple iterations, it was observed that Mixtral consistently produced similar elements regardless of the input document. This indicated structural similarities among the documents, allowing the newly generated elements to be reused across them, effectively identifying common patterns.
Once the ontology was finalized, a thorough manual review was conducted to eliminate superfluous elements in the schema. This step addressed inconsistencies and redundancies in \textit{DataProperty} elements, ensuring the ontology aligned with the project's requirements.\\
\textbf{- KG Creation:} Building on the created ontology and the RAG for each document, a prompt engineering phase was undertaken to generate KGs for each document. This phase utilized a few-shot prompting technique to design effective prompts. Once the individual KGs were generated, they were merged into a unified graph. The merging process preserved the distinct entities from each document and ensured consistency across the combined KG.\\
\textbf{- CQ's creation and Answering:} As a final step, CQs were generated using Mixtral. For this task, the ontology was provided to the LLM with instructions to create a list of CQs based on its structure. The generated CQs aligned seamlessly with the ontology schema, eliminating the need for manual refinement. However, only a subset of the CQs was selected for further use.
Subsequently, using Mixtral, a zero-shot prompt was developed to generate answers to the selected CQs based on each document. Each answer underwent a manual verification process to ensure consistency with both the reference KG and the original text, thereby avoiding potential hallucinations. Once validated, the verified responses were formatted into a CSV file for further analysis or use.

\item \textbf{NLP's Pipeline}

\textbf{- Preprocessing:} This step, as described above, was conducted to evaluate and compare new LLM-based approaches. The process was applied to only one of the five previously analyzed documents. Initially, after the document was read and its text extracted, a preprocessing phase was carried out using the NLTK library \footnote{\url{https://www.nltk.org/}}. This phase involved several steps, including punctuation removal, tokenization, stopword removal, and lemmatization, to prepare the text for subsequent analysis.\\
\textbf{- Part of Speech:} After tokenization, the SpaCy library was used to perform POS tagging for each token. POS tagging is a NLP technique that categorizes the components of a sentence, making it particularly valuable for complex tasks like Sentiment Analysis or Machine Translation. In this study, POS tagging was utilized to extract triples from the document. Tokens were categorized into "subj" (subject), "verb," and "obj" (object) based on their grammatical roles, enabling the creation of structured triples.\\
\textbf{- Triples Creation:} In the final step of the pipeline, triples were constructed by associating each identified subject with its corresponding verb and object. This process resulted in a structured representation of the text, capturing meaningful relationships between entities within the document.

\end{enumerate}
\begin{table}[t]
\centering
\caption{Score CQs answering of Large Language Models Approach.}\label{tab1}
\resizebox{0.7\textwidth}{!}{
\begin{tabular}{|l|c|c|}
\hline
\textbf{Results} &  \textbf{Full-text} & \textbf{Sub-part}\\
\hline
Which legal case is associated with Violation? & 3 & 4\\
\hline
What is the legal outcome of Case? & 4& 4\\
\hline
What is the reason stated for the judgment? &4 & 2\\
\hline
What abuse is related to Judgment? &  5& 4\\
\hline
What is the severity level of Abuse? &2 & 2\\
\hline
What is the duration and frequency of Abuse? &3& 2\\
\hline
Which legal articles are violated? & 4& 4\\
\hline
What is the context of Abuse?& 5& 4\\
\hline
Which court judged the Case? & 2& 1\\
\hline
What are the damages related to Judgment? &1& 3\\
\hline
What are the consequences of Abuse? &5&4\\
\hline
How much in legal damages and costs were awarded? & 0& 4\\
\hline
What is the legal status? & 5& 3\\
\hline
\multicolumn{1}{|r|}{\textbf{Total}} & \textbf{40/65}& \textbf{37/65}\\
\hline
\end{tabular}
}
\end{table}

\subsubsection{Empirical Evaluation}
To evaluate the LLM-based KG generation pipeline, we relied on a qualitative analysis based on answers to CQs. As detailed in Table 2, each of the five documents was associated with 13 CQs. These were answered by the LLM-generated KGs and manually verified against the source text. The full-text strategy yielded 40 consistent answers out of 65 (61.5\%), while the sub-part strategy provided 37 valid answers (56.9\%). This highlights a trade-off: full-text inputs offer broader coverage, while sub-part inputs although narrower yield more focused and relevant results, especially for factual or procedural queries. This CQ-based evaluation provided a useful lens into the structure, coverage, and correctness of the generated knowledge.

\subsubsection{Results}
This approach examines the role of LLMs in knowledge engineering tasks, including ontology generation, instance creation, and evaluation against traditional NLP methods. In generating ontologies, the LLM struggled to independently create a complete domain-specific ontology using only its pretrained knowledge, requiring additional domain-specific documents for specialization and highlighting the need for expert oversight in nuanced domains. Although the ontology generally performed well, the LLM occasionally failed to generate complete or accurate instances, even when tailored to the document structure, emphasizing the necessity of human validation to ensure correctness.

\begin{figure}[h]
\centering
\includegraphics[width=10cm]{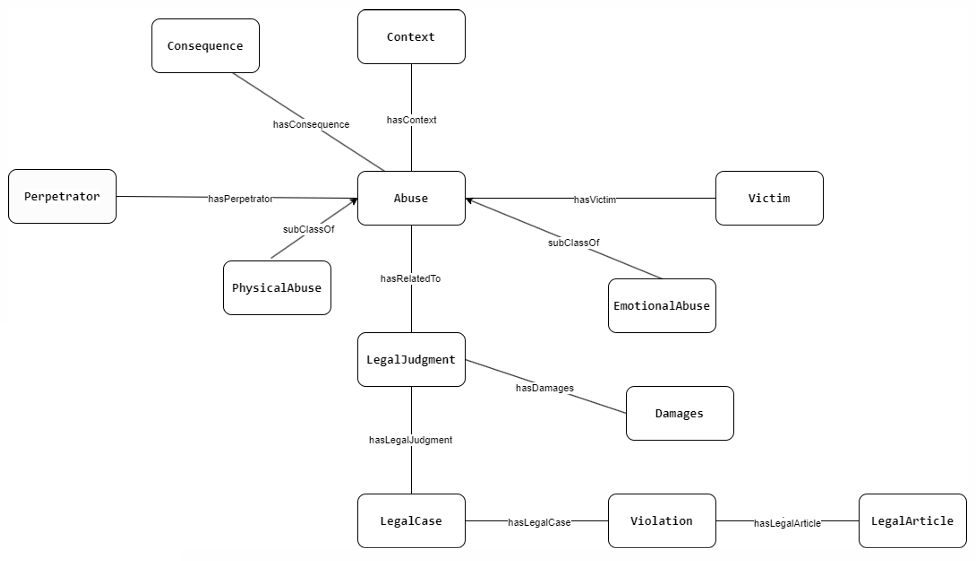}
\caption{Part of Knowledge Graph Constructed Using Large Language Models.}
\label{KGLLM}
\end{figure}

Two specific use cases were explored to evaluate the effectiveness of LLMs in document processing:\\
\textbf{\textit{Full-text approach:}} This use case involved processing the entire document to construct KGs and answer Competency Questions. While this approach provided the highest quantity of responses due to its comprehensive coverage, the relevance and accuracy of the responses varied. Full-text processing captured more nuanced information but required significant manual verification to filter out irrelevant or incorrect outputs. The technical challenges included longer processing times and token size limitations inherent in LLMs.\\
\textbf{\textit{Sub-part approach:}} This approach utilized text segments selected by domain experts, focusing on parts deemed most relevant to the analysis. Although this method produced fewer responses compared to the full text approach, it offered a higher likelihood of alignment with the objectives of the experts. It was faster, avoided token limitations, and facilitated targeted processing. However, it required precise expert input to ensure that no critical information was excluded.\\
The evaluation of the CQs, detailed in Table\ref{tab1}, focused on responses consistent with the text, excluding incomplete entities generated by the LLM only to establish links. The resulting KG, as defined in the GitHub repository \footnote{https://github.com/Fra3005/PreJust4Womans/blob/main/Commons/final\_onto.ttl},  represents structured knowledge derived from domain-specific legal and policy documents addressing women's rights and gender-based issues. It encapsulates core concepts, along with their semantic relationships comprising 12 classes, 9 object properties, 17 data properties, capturing the conceptual structure of the domain without populated example instances, as depicted in figure\ref{KGLLM}. This KG driven structure supports the instantiation of real-world scenarios discussed in source documents and serves as a foundation for answering CQs.\\
The CQs were classified according to relevance in the context of the study and their accuracy was confirmed by manual verification against the source documents. The comparison of these two use cases highlighted that while the full-text approach excels in information quantity, the sub-part approach proves more effective for specific, domain-oriented tasks. These findings underscore the importance of balancing comprehensive data processing with precision, guided by expert judgment, to maximize the utility and accuracy of LLM outputs.

\section{Discussion} \label{disc}

\subsection{Comparative Overview: Bottom-Up vs LLM-Based Approaches}
In this work we have proposed two distinct methodologies for constructing KGs to address legislation on violence against women: a bottom-up approach and an LLM-based approach. 
This section discusses the two approaches, comparing their methodologies, outputs, and suitability for our purpose. A summary of the discussion is provided in 
Tab.~\ref{tab:tab2}.

\begin{table}[h]
\centering
\caption{Comparative Analysis of Bottom-Up and LLM-Based Approaches.}
\resizebox{1\textwidth}{!}{%
\begin{tabular}{|p{3.5cm}|p{8cm}|p{8cm}|}\hline
\textbf{Criteria} &  \textbf{Bottom-Up Approach} & \textbf{LLM-Based Approach} \\ \hline
Methodology & Structured pipeline & Combine LLM capabilities with NLP and RAG techniques \\ \hline
Strengths& High precision and semantic alignment; Reliable for domain-specific; Structured tasks & Fast and scalable KG creation\\ \hline
Challenges& Labor-intensive and time consuming; less adaptable to novel or unexpected patterns  & Accuracy issues (hallucination and token limits \\ \hline
Performance on CQs &Excels in precise, structured queries using semantic standards  & Generate broader responses but needs post-processing to ensure relevance\\ \hline
Use Case Suitability & Best for formal legal reasoning and SPARQL-based tasks & Ideal for exploratory tasks, large scale processing, and rapid prototyping\\ \hline
Scalability & Limited scalability due to manual data processing & Highly scalable for diverse and large datasets \\ \hline
Output Consistency  & Outputs are precise and domain-aligned  & Outputs may vary in consistency and require refinement \\ \hline
Ontology Adaptability & Limited to predefined ontologies & Highly adaptable to diverse and nuanced data patterns \\ \hline
Ontology Concepts & Higher initially, with many specific concepts emerging from data & More limited, well-organized in a clear hierarchy \\ \hline
Ontology Relations & More dynamic, based on frequently occurring relations in data & Accurate and based on logical modeling \\ \hline
Ontology Size & Large in size (583.7 KB), including a more extensive set of classes, properties, and relationships  & Smaller in size (6.4 KB), suggesting a more concise ontology with potentially fewer classes and relationships \\ \hline
\end{tabular}
} 
\label{tab:tab2}
\end{table}

The bottom-up approach 
follows a systematic pipeline (involving data collection, knowledge extraction, triple integration, and ontology creation), ensuring the development of 
a KG, tailored for ECHR judgments, that is aligned and interlinked to existing resources, e.g. ECLI and Wikidata. 
This methodology resulted appropriate for delivering rather precise, domain-specific output, making it ideal for tasks requiring strict adherence to the available sources of knowledge and to existing semantically related resources. However, it is resource intensive, time-consuming, and less adaptable 
due to its reliance on predefined ontologies. In contrast, the LLM-based approach resulted a bit less structured in its process but rather flexible, leveraging advanced natural language processing and prompting techniques, such as zero-shot and few-shot prompting and retrieval-augmented generation, to accelerate ontology and KG creation. This approach is more scalable (given trained LLMs), adaptable to various legal texts, and capable of capturing nuanced information from full-text and expert selection document parts. Nevertheless, it faces challenges in accuracy and consistency, requiring extensive manual validation to avoid issues like hallucination and incomplete entities. While the bottom-up approach is suited for applications requiring precision, consistency, and semantic accuracy, such as formal legal reasoning and SPARQL-based querying, the LLM-based approach is more suitable for exploratory tasks, rapid prototyping, and large-scale data processing, where speed and adaptability are prioritized over perfect accuracy. 


\subsection{Ontology Development}
The project resulted in two ontologies; one generated using a Large Language Model, and one through a bottom-up approach. The automatic generated ontology, while efficient and scalable, was minimal and included generic classes due to the absence of domain specific constraints. In contrast, The manually constructed ontology was more semantically rich and aligned with legal reasoning needs. It introduced targeted classes to address specific gaps and accurately reflect legal structures. This comparison underscores the complementary value of both approaches and highlights the necessity of expert refinement when high legal fidelity is required. 

The demonstrated feasibility of combining structured extraction and LLM-based techniques opens new possibilities for semi-automated legal knowledge engineering pipelines. This hybrid approach could inform and the design of scalable KG generation frameworks across other legal domains, thereby contributing to the democratization of legal data access and interpretability in predictive justice systems. These findings underscore the importance of balancing comprehensive data processing with precision, guided by expert judgment, to maximize the utility and accuracy of LLM outputs. 

\subsection{Ensuring FAIRness in Legal Knowledge Engineering}
The developed Legal Knowledge Graph adheres to the FAIR (Findable, Accessible, Interoperable, and Reusable) principles through the following mechanisms:
\begin{itemize}
\item \textbf{Findable:} The KG is assigned a persistent DOI (https://doi.org/10.5281/zenodo.15270173) and is registered in the LOD Cloud (https://lod-cloud.net/dataset/PREJUST4WOMAN\_PROJECT), ensuring long-term retrievability and discoverability.
\item \textbf{Accessible:} The resource is openly accessible via Zenodo and GitHub, with no access restrictions. The KG is also exposed through a public SPARQL endpoint enabling both human and machine access.
\item \textbf{Interoperable:} The KG is published in standard RDF/Turtle format and relies on widely adopted vocabularies and ontologies, allowing integration with other legal and governmental data.
\item \textbf{Reusable:} The dataset is released under a permissive Creative Commons Attribution 4.0 International License, with complete documentation and source code available via GitHub. This enables others to reuse, extend, and adapt the resource for legal AI applications or broader knowledge engineering tasks.
\end{itemize}

\subsection{Quantitative Evaluation of Precision and Quality}
The evaluation provides a clearer understanding of the differences and complementarities between the two methodologies. The bottom-up approach emphasizes semantic precision, relying on structured extraction pipelines, curated ontologies, and expert-defined competency questions. This results in well-aligned, interpretable outputs that are suitable for formal legal reasoning and SPARQL querying. On the other hand, the LLM-based approach offers higher adaptability and faster development by leveraging retrieval-augmented generation and prompt engineering. While the automatically generated outputs require human oversight to ensure relevance and avoid hallucinations, they demonstrate strong potential for rapid KG bootstrapping and domain exploration. Competency questions were employed in both approaches: in the bottom-up pipeline, they guided ontology design and validation; in the LLM pipeline, they served as a mechanism to test the informativeness of generated graphs. Overall, the two approaches are complementary—manual engineering ensures domain fidelity and interpretability, while LLM-based methods enable scalability and prototyping across diverse legal texts.

\section{Conclusion}\label{conc}
With this paper, we 
introduced a novel 
Legal Knowledge Graph focused on legislation addressing violence against women. 
Designed in accordance with FAIR principles, the Legal KG ensures interoperability and possibly reusability across legal and AI systems. 
To construct the KG, two automated methodologies have been adopted/tailored: a 
bottom-up approach and an 
LLM-based approach. Each methodology offers distinct advantages: precision and domain-specific accuracy for the case of the bottom-up solution; scalability and flexibility for case of the LLM-based approach. 

Future work will focus on merging the two KGs into a unified resource, to be possibly expanded and on extending the KG applicability 
to the automated detection of legal patterns. 
Additionally, we are planning to experimenting the applicability of the proposed methodologies 
to other legal domains in order to showcase the generality of the proposed solutions and ultimately contributing to a unified, FAIR-aligned representation of European laws that may enhance 
legal automation and judicial efficiency.

\begin{credits}
    \subsubsection{\ackname}
    This work was partially supported by project \emph{FAIR - Future AI Research} (PE00000013), spoke 6 - Symbiotic AI (\url{https://future-ai-research.it/}) under the PNRR MUR program funded by the European Union - NextGenerationEU, and by PRIN project \emph{HypeKG - Hybrid Prediction and Explanation with Knowledge Graphs} (Prot. 2022Y34XNM, CUP H53D23003700006) under the PNRR MUR program funded by the European Union - NextGenerationEU
\end{credits}
%
%
%

\bibliographystyle{splncs04}
\bibliography{biblio}

\end{document}